%% file: jobresqa.tex
\documentclass[10pt, twocolumn]{article}

\usepackage{arxiv}

\usepackage{natbib}
\setcitestyle{authoryear,open={((},close={))}} 

\usepackage{todonotes}
\usepackage{tcolorbox}
\usepackage{adjustbox}
\usepackage{graphicx}
\usepackage{placeins}
\usepackage[utf8]{inputenc} 
\usepackage[T1]{fontenc} 
\usepackage{hyperref} 
\usepackage{url} 
\usepackage{booktabs} 
\usepackage{amsfonts} 
\usepackage{nicefrac} 
\usepackage{microtype} 
\usepackage{amsmath, amssymb}
\usepackage{graphicx}
\usepackage{hyperref}
\usepackage{algorithm}
\usepackage{algorithmic}
\usepackage{authblk}

\tcbuselibrary{listings,breakable}

\input{paul_tol_color_palette.tex}

\graphicspath{{figures/}}

\title{JobResQA: A Benchmark for LLM Machine Reading Comprehension on Multilingual
Résumés and JDs}

\author[1,2]{Casimiro Pio Carrino}
\author[1]{Paula Estrella}
\author[1]{Rabih Zbib}
\author[2]{Carlos Escolano}
\author[2]{José A. R. Fonollosa}

\affil[1]{Avature Machine Learning \\
\texttt{casimiro.carrino@avature.net} \\
\texttt{paula.estrella@avature.net} \\
\texttt{rabih.zbib@avature.net}
}

\affil[2]{Universitat Politècnica de Catalunya \\
\texttt{casimiro.pio.carrino@upc.edu} \\
\texttt{carlos.escolano@upc.edu} \\
\texttt{jose.fonollosa@upc.edu}
}

\begin{document}
    \twocolumn[
    \maketitle
    \begin{abstract}
        We introduce JobResQA, a multilingual Question Answering benchmark for
        evaluating Machine Reading Comprehension (MRC) capabilities of LLMs on
        HR-specific tasks involving résumés and job descriptions. The dataset comprises
        581 QA pairs across 105 synthetic résumé-job description pairs in five languages
        (English, Spanish, Italian, German, and Chinese), with questions
        spanning three complexity levels from basic factual extraction to
        complex cross-document reasoning. We propose a data generation pipeline derived
        from real-world sources through de-identification and data synthesis to ensure
        both realism and privacy, while controlled demographic and professional
        attributes (implemented via placeholders) enable systematic bias and
        fairness studies. We also present a cost-effective, human-in-the-loop
        translation pipeline based on the TEaR methodology, incorporating MQM error
        annotations and selective post-editing to ensure an high-quality multi-way
        parallel benchmark. We provide a baseline evaluations across multiple
        open-weight LLM families using an LLM-as-judge approach revealing higher
        performances on English and Spanish but substantial degradation for
        other languages, highlighting critical gaps in multilingual MRC capabilities
        for HR applications. JobResQA provides a reproducible benchmark for
        advancing fair and reliable LLM-based HR systems. The benchmark is publicly
        available at \url{https://github.com/Avature/jobresqa-benchmark}.
    \end{abstract}

    \vspace{1em}
    ]

    \section{Introduction}
    \label{sec:intro_new_version}

    The Human Resource (HR) field is undergoing a radical transformation with the
    use of AI. Large Language Models (LLMs) are increasingly being applied to
    tasks such as résumé parsing, candidate-job matching, interview evaluation, and
    conversational support. These models already outperform traditional keyword-based
    systems in candidate matching \cite{electronics14040794}, while HR-focused dialogue
    datasets demonstrate the potential of conversational HR agents \cite{xu-etal-2024-hr}.

    However, this rapid adoption raises significant concerns about accuracy, fairness,
    bias, and reproducibility. Controlled experiments show that current models
    often perpetuate demographic and cultural biases
    \cite{nghiem-etal-2024-gotta,rao2025invisiblefiltersculturalbias}. Such biases
    pose practical, ethical and even legal challenges under emerging frameworks
    such as the \emph{EU AI Act}\footnote{\url{https://artificialintelligenceact.eu/the-act/}},
    which emphasizes fairness, explainability, and responsible use.

    In light of these significant risks, there is an urgent need for benchmarks
    to evaluate LLM performance, fairness, and bias across HR tasks. Surveys on the
    use of NLP in the HR domain emphasize that reproducible and publicly available
    datasets are essential for assessing model consistency and transparency,
    especially in multilingual contexts \cite{otani-etal-2025-natural}. Recent works
    have started addressing these challenges by developing annotated datasets for
    skills and job matching \cite{gasco2025overview,zhang-etal-2022-skillspan} and
    by exploiting LLMs to generate synthetic résumés and job descriptions (JDs) that
    preserve privacy while enabling fairness studies \cite{fi15110363,saldivar2025syntheticcvsbuildtest}.

    One important use case of LLMs in HR is the analysis of résumés for matching
    with JDs. This task involves asking questions about the skills, experience,
    and background of a candidate in relation to a JD. Framing this process as a
    Machine Reading Comprehension (MRC) task enables knowledge-intensive Question
    Answering (QA) approaches that can better assess LLMs' reasoning about
    candidate-job suitability. While a few works have introduced HR-related QA datasets
    \cite{xu-etal-2024-hr,LuoLPG23,info13110513}, existing resources either
    focus on extractive, single-document CV questions or lack realistic,
    multilingual, and bias-controllable résumé-JD QA pairs.

    Motivated by these challenges, we introduce JobResQA, a synthetic, realistic,
    multilingual, and bias-controllable QA benchmark designed for recruiter-style
    questions over résumé-JD pairs. The dataset is derived from real-world data through
    a de-identification and synthesis pipeline, resulting in anonymized yet
    realistic résumés and JDs. JobResQA spans question types from basic factual extraction
    to complex, cross-document reasoning, and includes controlled demographic attributes
    for fairness analysis. It is annotated in English and extended to Spanish, Italian,
    German, and Chinese using a human-in-the-loop LLM translation pipeline.

    Our contributions are as follows:
    \begin{itemize}
        \item We release JobResQA: A curated, synthetic, multilingual QA dataset
            of over 105 résumé-JD pairs (581 QA items). It supports both short and
            long answers across three complexity levels: basic (extractive), intermediate
            (multi-passage), and complex (cross-document reasoning).

        \item We present a cost-effective, human-in-the-loop LLM translation
            pipeline that incorporates human feedback for quality estimation and
            refinement and produces high-quality parallel data in Spanish, Italian,
            German, and Chinese.

        \item We establish, to the best of our knowledge, the first comprehensive
            multilingual evaluation baseline to assess LLM's machine reading comprehension
            ability on résumés and JDs across several open-weight model families.
    \end{itemize}

    \section{Related Works}
    \label{sec:related_works} We group related research into three main areas.
    QA and MRC tasks in HR have been explored by \citet{xu-etal-2024-hr} with HR-MultiWOZ,
    the first HR-focused dialogue dataset, and \citet{LuoLPG23} who modeled
    résumé understanding as multilingual MRC by generating QA pairs from English
    and Dutch résumés.

    Synthetic data generation has proven effective for addressing data scarcity,
    with \citet{fi15110363} showing that ChatGPT-generated résumés improve job
    classification, while \citet{lorincz-etal-2022-transfer} and \citet{yu-etal-2025-confit}
    advanced vacancy generation and résumé matching through transfer learning and
    hypothetical embeddings.

    Bias and fairness research has identified critical issues, as \citet{saldivar2025syntheticcvsbuildtest}
    introduced demographic attributes in synthetic CVs for bias evaluation,
    \citet{nghiem-etal-2024-gotta} revealed name-based and gender biases in LLM employment
    recommendations, and \citet{rao2025invisiblefiltersculturalbias} exposed cultural
    biases in interview evaluations.

    These research directions inform JobResQA's design, which integrates QA
    framing, synthetic multilingual generation, and attribute-controlled data to
    enable comprehensive and fair evaluation of LLMs in HR applications.

    \section{The JobResQA Dataset}
    JobResQA is a QA benchmark over synthetic résumé-JD pairs designed to
    evaluate LLMs' machine reading comprehension for HR-specific applications.
    The dataset contains 581 question-answer (QA) pairs annotated over a set of
    105 unique pairs of résumé and JD. The résumés and JDs are derived from real-world
    data through a data synthesis pipeline that produces synthetic, anonymized, yet
    realistic versions (see Section \ref{sec:data_synthesis_pipeline}). The QA
    pairs are annotated manually by two linguists following detailed guidelines
    to ensure quality and diversity (see Section \ref{sec:qa_annotation}). The entire
    dataset is multi-way parallel across five languages: English (en), Spanish (es),
    Italian (it), German (de), and Chinese. The translations are produced by an LLM-based
    pipeline with human-in-the-loop corrections (see Section
    \ref{sec:translation_pipeline}). The set enables the evaluation of models in
    multilingual and cross-lingual QA setting, with questions in one language and
    résumé, JD in another language.

    \subsection{Main Characteristics}
    We designed the JobResQA benchmark to be realistic and representative of
    practical HR applications, capturing the complexity of real persons' career-related
    information and job requirements. We preserve the data's privacy and anonymity,
    while at the same time, we ensured certain properties to enable controlled studies
    in multilingual and fairness settings, as detailed below. The synthetic résumés
    and JDs are gender-inclusive and contain placeholders that represent controlled
    attributes across multiple bias dimensions (demographic, socioeconomic,
    educational, etc.), enabling systematic investigation of fairness in HR
    applications. In Table \ref{tab:controlled_attributes} we show a categorization
    of the placeholders extracted from the English data\footnote{For the other
    languages, we provide a translation dictionary to ensure parallelism, as
    described in Section \ref{sec:data_synthesis_review_and_post_processing}.}, along
    with bias-related dimensions they might impact.

    \begin{table}[!ht]
        \centering
        \small
        \begin{tabular}{l|l}
            \hline
            \textbf{Statistic}          & \textbf{Value}     \\
            \hline
            QAs (\#)                    & 581                \\
            Unique résumés (\#)         & 105                \\
            Unique JDs (\#)             & 101                \\
            Unique résumé-JD pairs (\#) & 105                \\
            QAs per résumé-JD pair (\#) & 5.53               \\
            \hline
            Complexity levels (\%)       \\
            \hspace{1mm} - Basic        & 26.5               \\
            \hspace{1mm} - intermediate & 36.7               \\
            \hspace{1mm} - Complex      & 36.8               \\
            \hline
            Languages supported         & en, es, de, it, zh \\
            \hline
        \end{tabular}
        \caption{JobResQA dataset statistics.}
        \label{tab:dataset_statistics}
    \end{table}
    \subsection{Statistics and Data Fields}

    \begin{table*}
        [!ht]
        \centering
        \begin{adjustbox}
            {max width=\textwidth}
            \begin{tabular}{p{4cm}p{8cm}p{4cm}}
                \hline
                \textbf{Attribute Category}                 & \textbf{ Placeholders}                                                                                                                     & \textbf{Bias Dimension}                                    \\
                \hline
                PII: Identity / Personal                    & \texttt{[NAME], [LASTNAME], [BIRTHDATE], [BIRTHPLACE], [NATIONALITY], [MARITAL\_STATUS]}                                                   & Demographic biases (gender, ethnicity, nationality)        \\
                \hline
                PII: Contact / Location                     & \texttt{[ADDRESS], [APARTMENT], [CITY], [STATE], [STATES], [ZIPCODE], [COUNTRY], [COUNTRIES], [LOCATION], [LOCATIONS], [AIR\_FORCE\_BASE]} & Geographic or socioeconomic biases                         \\
                \hline
                PII: Online Profiles                        & \texttt{[EMAIL], [PHONE], [SKYPE\_ID], [LINKEDIN\_PROFILE], [GITHUB\_URL]}                                                                 & Identification / privacy risk (not directly bias-relevant) \\
                \hline
                Affiliation: Education                      & \texttt{[UNIVERSITY], [COLLEGE], [SCHOOL], [INSTITUTION]}                                                                                  & Prestige or socio-educational biases                       \\
                \hline
                Affiliation: Employment / Organization      & \texttt{[COMPANY], [PARENT\_COMPANY], [ORGANIZATION], [INDUSTRY], [COMPANY\_WEBSITE]}                                                      & Domain or prestige bias                                    \\
                \hline
                Professional Context: Job / Role            & \texttt{[POSITION], [WORK\_STATUS], [TEAM], [SUPERVISOR], [PLATFORM]}                                                                      & Core job semantics (usually neutral)                       \\
                \hline
                Qualification: Professional Credentials     & \texttt{[CERTIFICATION], [LICENSE], [AWARD], [PRODUCT], [SECURITY\_CLEARANCE\_STATUS]}                                                     & Domain, prestige bias or sensitive (clearance level)       \\
                \hline
                Temporal Identifiers                        & \texttt{[DATE], [DEADLINE]}                                                                                                                & Neutral / temporal only                                    \\
                \hline
                Professional Context: Project / Client Info & \texttt{[CLIENT], [CLIENTS], [DETAILS]}                                                                                                    & Prestige or domain bias                                    \\
                \hline
            \end{tabular}
        \end{adjustbox}
        \caption{Controlled attribute categories, example placeholders, and
        associated bias dimensions. Multilingual versions use translated placeholders
        to maintain parallelism.}
        \label{tab:controlled_attributes}
    \end{table*}

    We report the main statistics of the JobResQA benchmark in Table \ref{tab:dataset_statistics}
    and describe briefly the main dataset's textual fields\footnote{For brevity,
    we omit containing numerical identifiers} as below:
    \begin{itemize}


        \item \texttt{resume}: text of synthetic candidate's résumé.


        \item \texttt{jd}: synthetic description of a role.


        \item \texttt{question}: recruiter-style questions on the résumé in relation
            to the JD.

        \item \texttt{short\_answer}: concise answer to the question, as a span,
            phrase, number, or yes/no.

        \item \texttt{explanation} for the provided short answer, typically a longer
            answer with few sentences containing explanatory rationale and evidences.


        \item \texttt{complexity level} (of the question): basic (extractive-style),
            intermediate (knowledge-intensive), complex (multi-hop, cross-document
            reasoning).

        \item \texttt{language}: language of all text fields.



    \end{itemize}

    \subsection{Accessibility and Reproducibility}
    \label{sec:accessibility_reproducibility}
    We release JobResQA under the Creative Commons BY-SA 2.0 (Attribution-ShareAlike
    2.0 Generic) license. The dataset is available at our official GitHub repository\footnote{\url{https://github.com/Avature/jobresqa-benchmark}},
    which includes comprehensive documentation, usage instructions, and supplementary materials.
    These materials comprise all prompts used to instruct the LLMs for dataset synthesis
    and translation, the MQM error categories with annotated examples, placeholder
    lists and translation dictionaries, QA annotation guidelines, and additional resources.

    \section{Résumés and Job Synthesis}
    \label{sec:data_synthesis_pipeline} In this section, we describe how the JobResQA
    dataset were created, detailing the generation of synthetic, anonymized résumé-JD
    matched pairs and the QA annotation to create recruiter-style questions and answers
    intended to assess the machine reading comprehension capabilities of LLMs.

\begin{figure}[!h]
        \centering
        \includegraphics[width=\columnwidth]{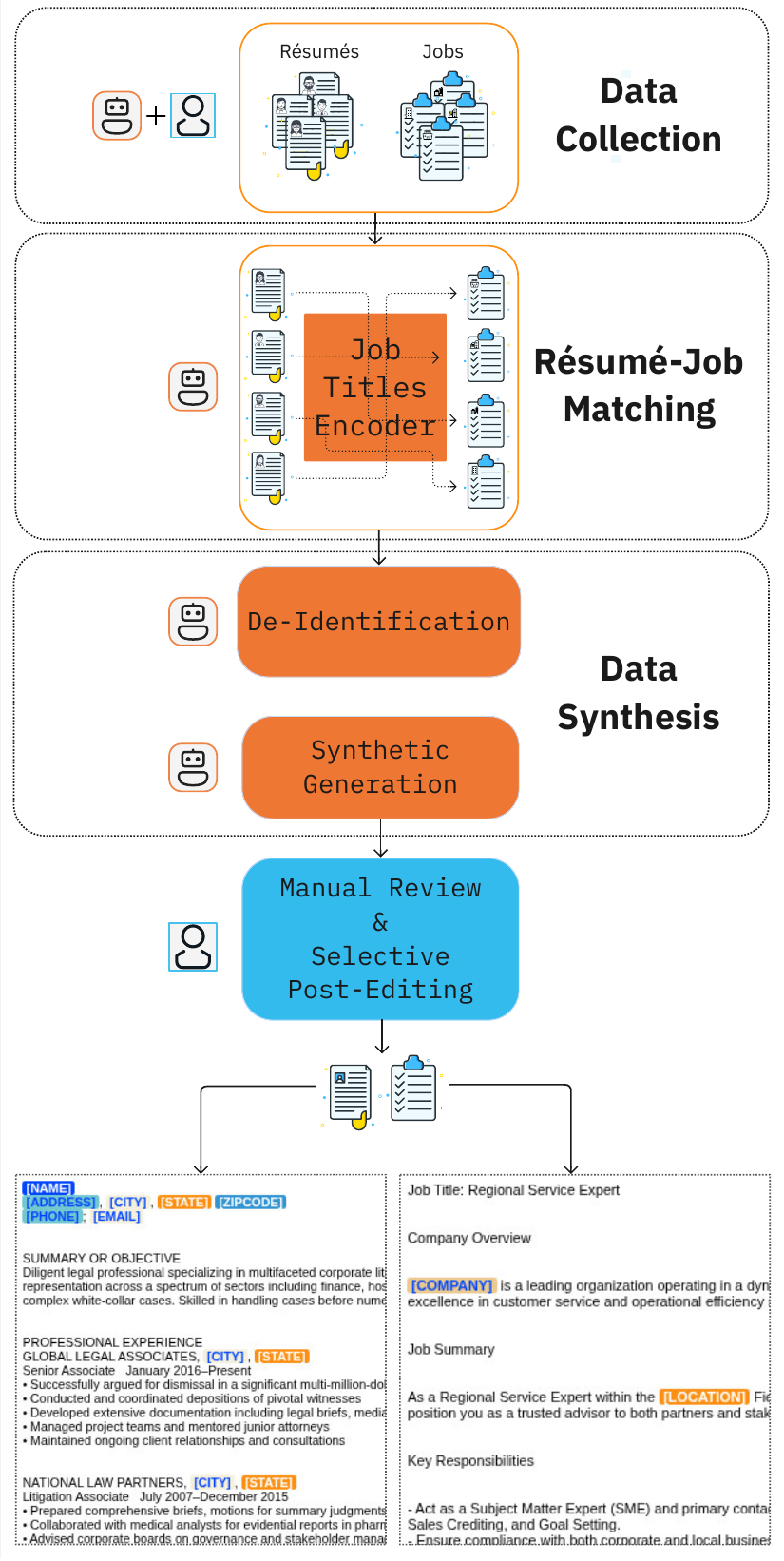}
        \caption{Data synthesis pipeline for JobResQA: collection and matching of
        résumé-JD pairs, de-identification, LLM-based synthesis, manual review
        and post-processing.}

        \label{fig:data_synthesis_pipeline}
    \end{figure}

    The process involves multiple stages that we illustrate in Figure \ref{fig:data_synthesis_pipeline}:
    data collection, résume-JD matching, de-identification and synthetic
    generation and a final manual review and post-processing for better quality
    and consistency.

    \begin{figure}[!ht]
        \centering
        \includegraphics[width=\columnwidth]{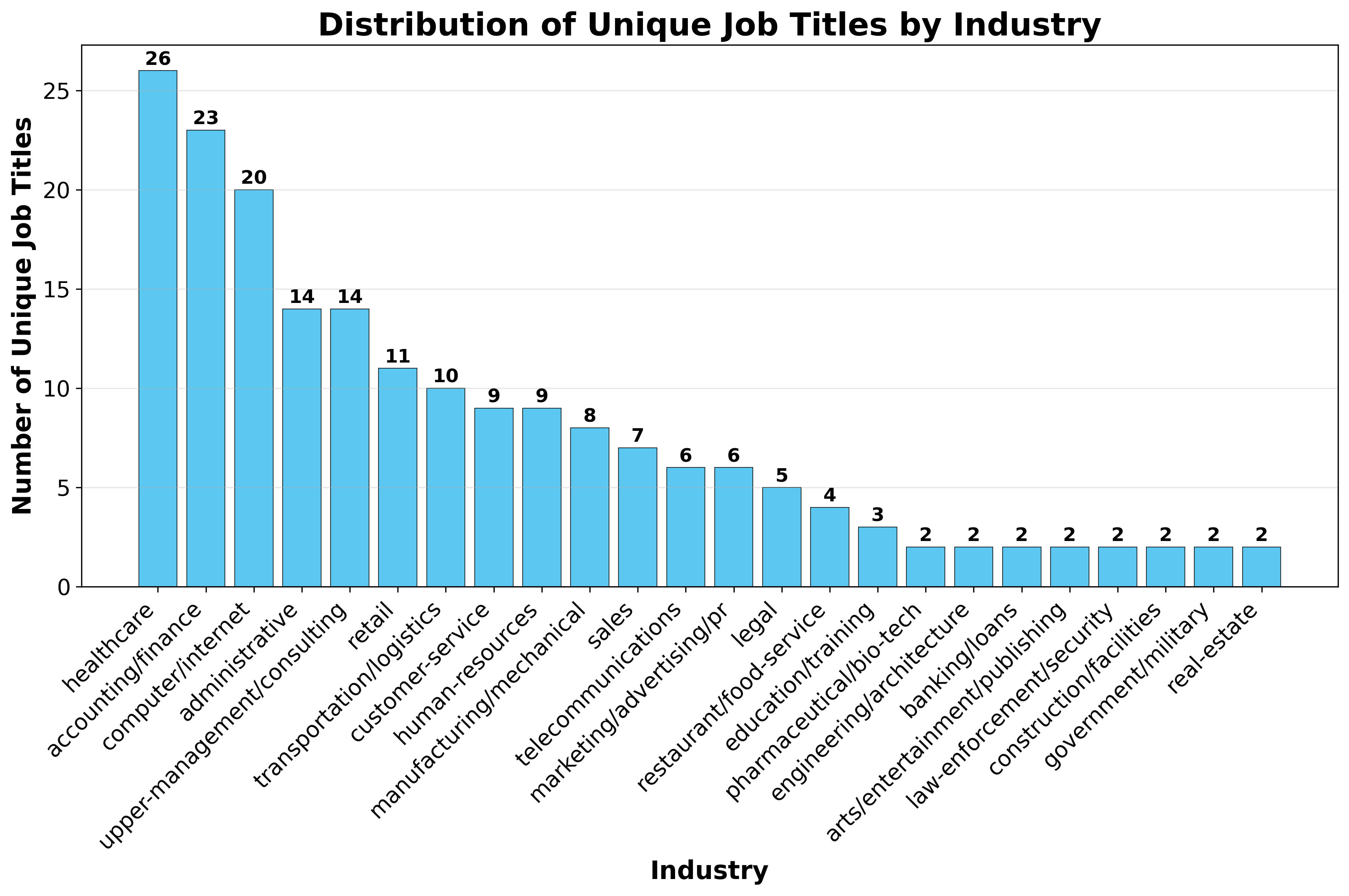}
        \caption{Industry distribution of the final selection of matched 105
        résumé-JD pairs, with 24 industries and 194 unique job titles.}
        \label{fig:industry_distribution}
    \end{figure}

    \subsection{Data Collection and Job Matching}
    We start by collecting real-world résumés and JDs from a large pool of
    public job boards that are randomly sampled from diverse locations and
    industries to target a wide array of roles and domains. Then, we align
    candidates with suitable roles by performing semantic matching using the job
    titles of the résumés and JDs. We use the multilingual job title encoder in \citet{deniz2024jobtitle}
    to encode the job title of résumés and JDs into a shared embedding space, and
    compute cosine similarity to identify the most similar pairs. In particular,
    given a résumé job title, obtain the top-10 JDs titles from the ranking, and
    then we manually review and select the best match based on title similarity and
    industry alignment. The manual review improves the quality over automatic
    threshold-based selection alone.

    The final result is a selection of 105 matched résumé-JD pairs covering a
    total of 24 industries and 194 unique job titles, as shown in Figure \ref{fig:industry_distribution}.

    \subsection{De-identification}
    \label{sec:de-identification} We then pass the records through a de-identification
    stage to preserve the privacy of the data. For résumés we use the model in
    \citet{DBLP:conf/hr-recsys/RetykFATZ23} to extract relevant entities such as
    contact information, work experience, education, and languages, and we replace
    all but the job titles and skills with placeholders (\texttt{[NAME], [PHONE],
    etc.}).

    For JDs, we implement a rule-based de-identification approach by creating a list
    of companies, branches, products and company-related identifiable entities and
    then extracting and replacing those with placeholders (e.g., [COMPANY], [PRODUCT],
    etc.) to remove traceability to the original company.

    \subsection{Synthetic Generation}
    We generated synthetic versions from de-identified résumés and JDs that
    preserve original content while ensuring anonymization. We employ an LLM-based
    approach with carefully crafted prompts using OpenAI's GPT-4.1 with temperature
    set to 0.7 and \textit{top\_p} to 1.

    For résumé generation, we apply three key transformations: 1) anonymizing
    personal information by replacing all PII with placeholders; 2) modifying
    career details through career-aware changes that preserve professional
    meaning while obscuring identifiers, job titles and skills are replaced with
    pertinent alternatives, descriptions are rephrased, dates are shifted maintaining
    chronological consistency, and indirect identifiers are modified; (3) preserving
    format and style by mapping content to clear sectioning (Experience, Education,
    Skills) while maintaining comparable length and realistic flow.

    Similarly, JD generation involves: 1) anonymizing company information by
    replacing all company-related identifiable details with placeholders; 2)
    rephrasing job content to remove distinctive wording while preserving role-specific
    aspects including job title, skills, responsibilities, and requirements; 3) preserving
    format and style by maintaining comparable length, professional tone, and realistic
    formatting.

    \subsection{Manual Review and Selective Post-Editing}
    \label{sec:data_synthesis_review_and_post_processing} Finally, we manually
    reviewed the résumés and JDs by applying a selective post-editing that
    results in a high-quality of the anonymized texts. We targeted minor issues (e.g.,
    typos and formatting inconsistencies) and checked for any remaining personally
    identifiable information. We also replaced sentences directly referencing a
    gender with gender-neutral alternatives to promote inclusiveness (e.g., "the
    first female to" -> "the first person to"). Finally, we manually normalized all
    placeholders to be consist across the dataset (see Table
    \ref{tab:controlled_attributes}). This combined process of pipeline, de-identification,
    synthesis, manual review, and post-editing produces 105 synthetic résumé-JD
    pairs that preserve realistic professional content while ensuring anonymity.
    The resulting data set enables controlled fairness studies of demographic and
    professional attributes.

    \section{Question-Answering Annotations}
    \label{sec:qa_annotation}

    Two linguists annotated the dataset, focusing on diversity in question creation
    and writing style. Following prior work on QA resource development (\citet{sen-etal-2022-mintaka,yang-etal-2018-hotpotqa,10.1109/TKDE.2022.3223858}),
    we first conducted a pilot study on a small set of résumé-JD pairs. Insights
    from this pilot informed the design of refined annotation guidelines. During
    the main annotation phase, the linguists created triplets of \textit{(question,
    short answer, explanation)} focusing on specific candidate aspects (e.g., work
    experience). Each triplet was assigned one of three complexity levels:

    \begin{itemize}
        \item \textbf{Basic}: Single-passage questions referring primarily to résumé
            content, optionally involving JD information. For example, the
            question \textit{“Does the candidate meet the basic educational
            requirements?”} requires comparing the education sections from the résumé
            and the JD.

        \item \textbf{Intermediate}: Questions requiring reasoning across multiple
            résumé and JD sections. For example, the question \textit{“How
            aligned is the candidate's experience with the stakeholder
            collaboration requirements?”}, involves analyzing work experience,
            skills, and evidence of collaboration abilities across résumé sections.

        \item \textbf{Complex}: Questions demanding deeper reasoning, external
            knowledge integration, or interpretation beyond direct fact lookup.
            For example, the question \textit{“What value does bioinformatics
            research bring to document management systems?”}, requires inferring
            the transferability of domain-specific skills.
    \end{itemize}

    For better interpretability and consistency, annotators provided both short answers
    and explanations detailing their reasoning process. When formulating
    questions, they were instructed to avoid targeting placeholders or gender-specific
    information in résumés and JDs, focusing instead on generalizable skills, experiences,
    and qualifications and other job and career-related aspects. For support,
    annotators used a curated question bank developed through consultancy
    sessions with Talent Acquisition professionals, to ensure alignment with real-world
    HR screening practices. Moreover, given the dataset’s broad industry coverage,
    the linguists had the opportunity to consult the ESCO dictionary \cite{esco}
    and the O*NET database \cite{ONET2025} to clarify unfamiliar job titles, skills,
    or domain-specific terminology.

    \section{Human-in-the-Loop Machine Translation Pipeline}
    \label{sec:translation_pipeline} To evaluate LLMs’ capabilities in HR-specific
    machine reading comprehension tasks across multiple languages, we extended JobResQA
    to four additional languages: Spanish, Italian, German, and Chinese.
    Building on recent studies showing that LLMs can produce high-quality
    translations in zero- and few-shot settings \cite{feng-etal-2025-tear, zhu-etal-2024-multilingual,
    cui-etal-2025-multilingual, koshkin-etal-2024-llms}, we developed an LLM-based
    machine translation pipeline with human-in-the-loop feedback.

\begin{figure}[!htbp]
        \centering
        \includegraphics[width=\columnwidth]{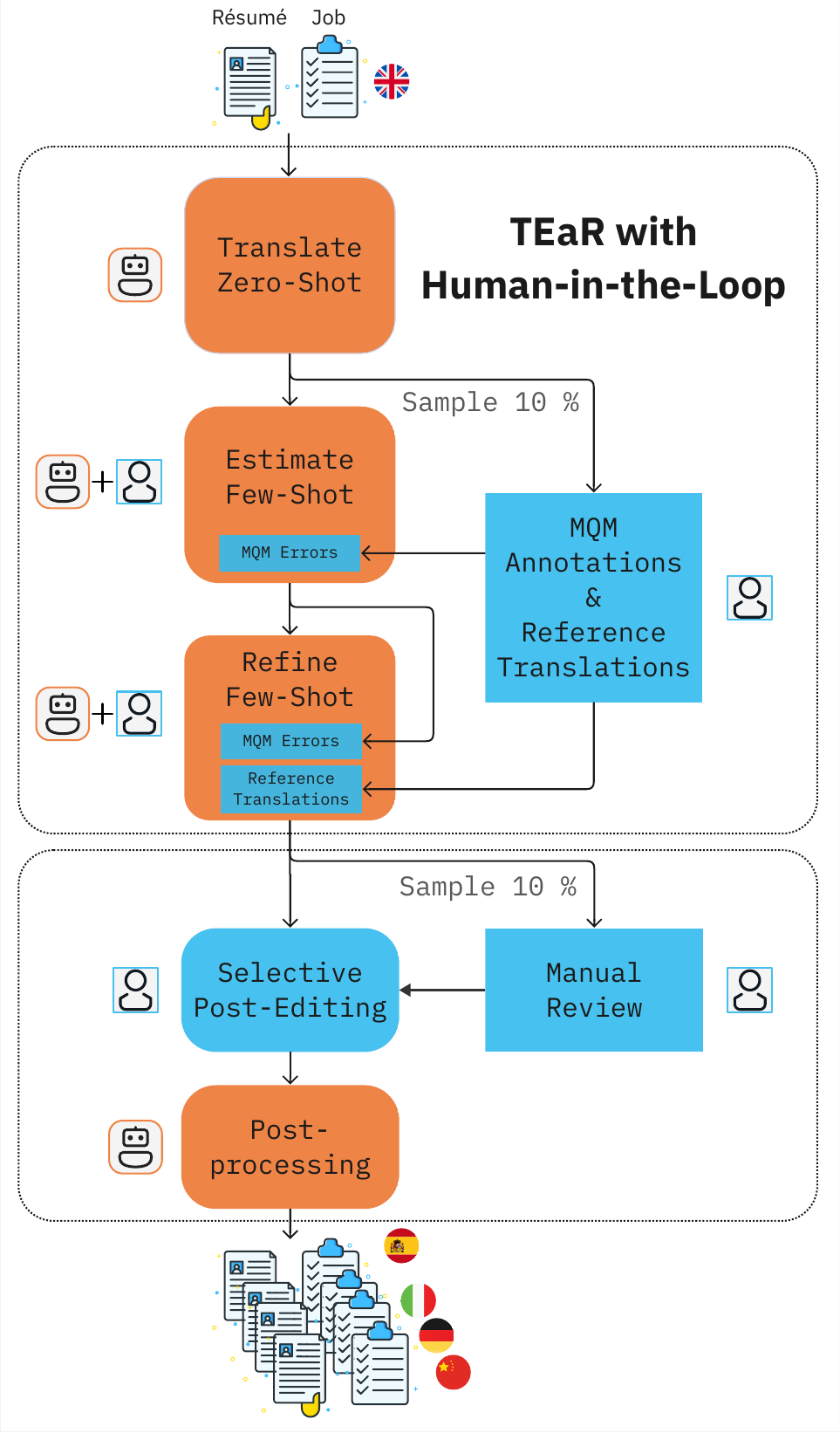}
        \caption{Human-in-the-loop TEaR translation pipeline for JobResQA: zero-shot
        translation, MQM error annotation and corrections, few-shot estimation,
        few-shot refinement and selective post-editing.}

        \label{fig:translation_pipeline}
    \end{figure}

    Our multi-stage translation process, designed specifically for résumés and job
    descriptions (JDs), includes machine translation, human error annotation,
    selective post-editing, and post-processing (Figure\ref{fig:translation_pipeline})
    . We combined automatic translation using \textit{Claude Sonnet 4}\footnote{anthropic.claude-sonnet-4-20250514-v1:0}
    (temperature = 0) with review and error annotation by native speakers, thus balancing
    translation quality with efficiency. We run inference using AWS Bedrock
    service\footnote{\url{https://aws.amazon.com/bedrock/}}

    \subsection{TEaR with Human-in-the-Loop}
    We implemented a human-in-the-loop variant of Translate-Estimate-Refine (TEaR)
    \cite{feng-etal-2025-tear} using MQM errors
    \cite{lommel2013multidimensional} and corrected translations for feedback
    during the Estimate and Refine steps. The steps of our TEaR-based human-in-the-loop
    translation pipeline are:

    \paragraph{Zero-Shot Translation.}
    We started with an initial translation following the zero-shot translation
    prompt strategy in \citet{feng-etal-2025-tear}, with an additional
    instruction that preserve placeholders and formatting. We translated résumés
    and JDs at the paragraph-level, while other fields were processed entirely. At
    this stage, the idea is to generate LLMs without any human guidance, which are
    later improved with human-in-the-loop feedbacks.

    \paragraph{Human Feedback: MQM Errors Annotations and Corrected Translations.}
    We sample ~10\% of the résumé-JD pairs for manual review. Annotators identify
    translation errors using our custom MQM categories designed for HR documents:
    Terminology, Accuracy, Linguistic, Style, Locale, Design, and Custom.

    Notably, we introduced {\em Hallucination} under the {\em Custom} category to
    capture AI-generated content errors, and {\em Gender-Inclusive} under the
    Style category to address concerns about inclusiveness in the translations. The
    gender-inclusive error category targets gender-specific translations and enforces
    a corrections that uses slashed forms (e.g., "des/der Kandidaten/-in", "el/la
    candidato/a", "del/la candidato/a"). All errors were rated across four severity
    levels: Critical, Major, Minor, and Neutral. This feedback guided the LLM in
    subsequent {\em Estimate and Refine} steps, driving it towards higher-quality
    translations and better aligned with human preferences. Table
    \ref{tab:mqm_errors_categories} shows the MQM error categories along with
    the severity levels used by human translators on the translated texts.
    \begin{table}[!ht]
        \centering
        \small
        \begin{tabular}{l|l}
            \hline
            \textbf{Category}   & \textbf{Subcategory}                 \\
            \hline
            Terminology         & Inconsistent Use, Wrong term         \\
            Accuracy            & Mistranslation, Addition, Omission   \\
            Linguistic          & Grammar, Punctuation, Spelling       \\
            Style               & Inconsistent style, Gender Inclusive \\
            Locale              & Entity Format                        \\
            Design              & Layout                               \\
            Custom              & Hallucination                        \\
            \hline
            \textbf{Severities} & Critical, Major, Minor, Neutral      \\
            \hline
        \end{tabular}
        \caption{MQM errors categories and Severity levels used for human
        annotations}
        \label{tab:mqm_errors_categories}
    \end{table}

    \paragraph{Few-Shot Estimation.}
    We utilized the errors from human MQM annotations to apply the few-shot estimation
    prompting strategy from \citet{feng-etal-2025-tear}. This allowed us to automatically
    scale error estimation to the entire dataset following the MQM error categories
    we defined. These errors provide feedback to improve subsequent translations.

    \paragraph{Few-Shot Refinement.}
    Finally, we fed both the estimated MQM errors and corrected translations (used
    as references) to apply the few-shot refinement prompting strategy in \citet{feng-etal-2025-tear}.
    Similar to the estimation step, this allows us to scale the refinement to the
    entire dataset. The corrected translations provide references that guide the
    LLM to refine the initial translations based on human feedback and
    preferences.

    \subsection{Manual Review, Selective Post-Editing, and Post-Processing.}
    To ensure high-quality translations, we sampled 10\% of translated résumé-JD
    pairs for manual review by native speakers to identify issues to be
    addressed through selective post-editing, either manually or automatically.
    Below, we describe the main issues found and how we addressed them:

    \paragraph{Job Titles Consistency.}
    We manually detected remaining untranslated English job titles and replaced
    them with target-language equivalents, ensuring consistency across all dataset
    fields.

    \paragraph{Automatic Verb Tenses Consistency.}
    For each résumé, we normalized verb-tense usage to avoid inconsistent shifts
    between tenses within work-experience sections. We applied an LLM-based post-editing
    step using \textit{Claude Sonnet 4}\footnote{anthropic.claude-sonnet-4-20250514-v1:0}
    (temperature = 0) to enforce deterministic rules with light rewriting where
    needed. The rules consisted of using present-tense verbs or nominalized forms

    for the candidate's current or most recent position, and nominalized forms for
    past positions. We also instructed the model to remove first-person pronouns
    to produce more natural résumé-style descriptions.

    \paragraph{Gender-Inclusive Forms.}
    For Spanish, Italian and German, we detected and fixed gender-inclusive form
    issues. We automatically extracted all words containing the gender-inclusive
    slash "/" using a rule-based approach (e.g., "des/der Kandidaten/-in", "el/la
    candidato/a", "del/la candidato/a"), then manually fixed each occurrence to
    ensure consistency with MQM annotated errors.

    \paragraph{Placeholders Normalization and Translations.}
    We created a dictionary of placeholder translations from English to target languages,
    replacing all placeholders in translated texts, thus maintaining consistency
    across the entire dataset.

    In Table \ref{tab:counts_edits}, we report total edit counts after the
    selective post-editing step split by language and dataset's field, with the résumé
    requiring the most edits overall.
    \begin{table}[!ht]
        \centering
        \small
        \begin{tabular}{l|c|c|c|c}
            \textbf{Field}         & \textbf{de} & \textbf{es} & \textbf{it} & \textbf{zh} \\
            \hline
            \texttt{resume}        & 2494        & 2776        & 2754        & 1821        \\
            \texttt{jd}            & 612         & 500         & 543         & 368         \\
            \texttt{short\_answer} & 267         & 137         & 278         & 315         \\
            \texttt{explanation}   & 142         & 62          & 110         & 23          \\
            \hline
        \end{tabular}
        \caption{Total edit counts from manual post-editing and automated post-processing.}
        \label{tab:counts_edits}
    \end{table}

    \subsection{Translation Quality Evaluation}
    To evaluate the quality of the final translations, we employed two widely-used
    reference-free metrics: COMETKiwi \cite{rei-etal-2022-cometkiwi,rei-etal-2023-scaling}
    and BLASER 2.0 \cite{dale-costa-jussa-2024-blaser}. Both metrics are
    specifically designed for quality estimation of machine translations when
    reference translations are unavailable.

    Table \ref{tab:translation_quality_estimation} presents the average scores, across
    all target languages, for the final translations and their improvements (delta)
    over the initial zero-shot translations from the {\em Translate} step. The results
    demonstrate consistently high translation quality across all languages, with
    COMETKiwi scores ranging from 83.36 to 85.51 and BLASER 2.0 scores ranging
    from 4.25 to 4.62. Furthermore, the positive delta scores for COMETKiwi
    metrics, from 0.05 to 0.45, confirm that the human-in-the-loop feedback
    combined with selective post-editing successfully improved translation
    quality across all languages.

    \begin{table}[!ht]
        \centering
        \small
        \begin{tabular}{l|c|c|c}
            \hline
            \textbf{Lang} & \textbf{COMETKiwi}     & \textbf{COMETKiwi}     & \textbf{BLASER 2.0} \\
                          & \textbf{(2022)}        & \textbf{(2023, XL)}    &                     \\
            \hline
            de            & 83.36 (\textbf{+0.09}) & 74.65 (\textbf{+0.16}) & 4.59 (-0.05)        \\
            es            & 85.25 (\textbf{+0.33}) & 78.08 (\textbf{+0.43}) & 4.62 (0.00)         \\
            it            & 85.51 (\textbf{+0.24}) & 78.95 (\textbf{+0.45}) & 4.61 (-0.02)        \\
            zh            & 83.07 (\textbf{+0.05}) & 74.86 (\textbf{+0.19}) & 4.25 (-0.02)        \\
            \hline
        \end{tabular}
        \caption{Translation quality estimation scores of the final translations
        and delta over the initial zero-shot translation from the Translate step.}
        \label{tab:translation_quality_estimation}
    \end{table}


    \section{Evaluation Experiments}
    The goal of this section is to establish a first baseline evaluation on the
    JobResQA benchmark to assess LLMs machine-reading comprehension through
    question answering on résumé and JDs across languages. In the following, we
    describe the experimental setup, including the models, prompting strategy
    and evaluation metrics, and then we discuss the results. We run inference
    using AWS Bedrock service\footnote{\url{https://aws.amazon.com/bedrock/}},
    which provides access to a variety of foundation models through API
    endpoints.

    \subsection{Experimental Setup}
    \label{sec:experimental_setup}

    \begin{table*}
        [!ht]
        \centering
        \begin{tabular}{l|c|c|c|c|c|c}
            \hline
            \textbf{Model}         & \textbf{Size} & \textbf{en}                                         & \textbf{es}                                         & \textbf{de}                                         & \textbf{it}                                         & \textbf{zh}                                         \\
            \hline
            Mistral Large (2402)   & large         & \cellcolor{TolSunset5!80}{0.69} $\pm$ 0.26          & \cellcolor{TolSunset5!80}{0.67} $\pm$ 0.25          & \cellcolor{TolSunset5!80}{\textbf{0.65}} $\pm$ 0.25 & \cellcolor{TolSunset5!80}{0.66} $\pm$ 0.24          & \cellcolor{TolSunset5!80}{0.61} $\pm$ 0.29          \\
            Mistral Small (2402)   & large         & \cellcolor{TolSunset5!80}{0.65} $\pm$ 0.28          & \cellcolor{TolSunset5!80}{0.61} $\pm$ 0.28          & \cellcolor{TolSunset7!80}{0.59} $\pm$ 0.29          & \cellcolor{TolSunset7!80}{0.60} $\pm$ 0.29          & \cellcolor{TolSunset7!80}{0.40} $\pm$ 0.28          \\
            Llama 3.3 70B Instruct & large         & \cellcolor{TolSunset5!80}{\textbf{0.73}} $\pm$ 0.26 & \cellcolor{TolSunset5!80}{\textbf{0.69}} $\pm$ 0.25 & \cellcolor{TolSunset7!80}{0.47} $\pm$ 0.38          & \cellcolor{TolSunset5!80}{\textbf{0.70}} $\pm$ 0.25 & \cellcolor{TolSunset7!80}{0.48} $\pm$ 0.39          \\
            Llama 3.1 70B Instruct & large         & \cellcolor{TolSunset5!80}{0.72} $\pm$ 0.26          & \cellcolor{TolSunset5!80}{0.68} $\pm$ 0.25          & \cellcolor{TolSunset5!80}{0.64} $\pm$ 0.27          & \cellcolor{TolSunset7!80}{0.43} $\pm$ 0.38          & \cellcolor{TolSunset5!80}{\textbf{0.66}} $\pm$ 0.27 \\
            \hline
            Llama 3.2 3B Instruct  & medium        & \cellcolor{TolSunset7!80}{0.55} $\pm$ 0.27          & \cellcolor{TolSunset7!80}{0.49} $\pm$ 0.26          & \cellcolor{TolSunset7!80}{0.45} $\pm$ 0.25          & \cellcolor{TolSunset7!80}{0.47} $\pm$ 0.26          & \cellcolor{TolSunset7!80}{0.47} $\pm$ 0.28          \\
            Llama 3.2 1B Instruct  & medium        & \cellcolor{TolSunset7!80}{0.35} $\pm$ 0.25          & \cellcolor{TolSunset9!80}{0.27} $\pm$ 0.24          & \cellcolor{TolSunset9!80}{0.25} $\pm$ 0.20          & \cellcolor{TolSunset9!80}{0.24} $\pm$ 0.20          & \cellcolor{TolSunset9!80}{0.26} $\pm$ 0.21          \\
            Llama 3.1 8B Instruct  & medium        & \cellcolor{TolSunset5!80}{0.62} $\pm$ 0.30          & \cellcolor{TolSunset7!80}{0.57} $\pm$ 0.29          & \cellcolor{TolSunset7!80}{0.52} $\pm$ 0.28          & \cellcolor{TolSunset7!80}{0.56} $\pm$ 0.29          & \cellcolor{TolSunset7!80}{0.56} $\pm$ 0.30          \\
            Gemma 3 4B Instruct    & medium        & \cellcolor{TolSunset5!80}{0.64} $\pm$ 0.29          & \cellcolor{TolSunset7!80}{0.39} $\pm$ 0.21          & \cellcolor{TolSunset7!80}{0.48} $\pm$ 0.28          & \cellcolor{TolSunset7!80}{0.40} $\pm$ 0.21          & \cellcolor{TolSunset7!80}{0.41} $\pm$ 0.22          \\
            Gemma 3 1B Instruct    & medium        & \cellcolor{TolSunset9!80}{0.29} $\pm$ 0.17          & \cellcolor{TolSunset9!80}{0.15} $\pm$ 0.11          & \cellcolor{TolSunset9!80}{0.15} $\pm$ 0.11          & \cellcolor{TolSunset9!80}{0.16} $\pm$ 0.11          & \cellcolor{TolSunset9!80}{0.15} $\pm$ 0.10          \\
            \hline
        \end{tabular}
        \caption{Average G-Eval scores (mean ± std) by model, size and language.
        The score ranges are based on evaluation rubrics: factually incorrect (0.0-0.3),
        mostly correct (0.3-0.6), correct but missing minor details (0.6-0.9).
        Higher scores indicate better alignment with human reference answers.}
        \label{tab:qa_eval_main_results}
    \end{table*}

    \paragraph{Performing QA with LLMs.}
    We designed a zero-shot prompt that instructs the model to act as an expert hiring
    assistant professional, answering questions about a candidate using only the
    JD and the provided résumé. Following the QA annotation guidelines in Section
    \ref{sec:qa_annotation}, the model is prompted to produce a concise short
    answer and a detailed explanation strictly grounded in the résume and/or JD.
    Responses should be factual, objective, and in the same language as the question,
    with explanations referencing specific details, quotes as evidence, and with
    justification for any information inferred from résumés and JDs.

    For the QA task, we experimented with several open-weight, multilingual LLM
    models from various families and sizes, from medium to large. The selected
    models are: Llama 3.1\cite{weerawardhena2025llama31foundationaisecurityllm8binstructtechnicalreport}
    and 3.2 (1B, 3B, 8B, 70B) , Mistral 2402 version (Small and Large), and
    Gemma 3 (1B and 4B) \cite{gemmateam2025gemma3technicalreport}. We set the temperature
    to 0, and maximum response length to 512 tokens to produce deterministic and
    concise outputs.

    \paragraph{LLM-as-a-Judge Evaluation}
    \label{sec:llm_as_judge} We evaluated model answers using an LLM-as-a-judge framework
    with the G-EVAL metric \cite{liu-etal-2023-g}, measuring alignment with
    human responses. Concretely, we provided a list of evaluation steps that
    guide the judge to compare the short answer and explanations from both the
    model and human responses. The judge checks that the short answer is concise
    and it uses minimal wording, and that the explanation provides detailed
    justification with specific references to the JD or résumé. The judge determines
    whether both answers communicate the same main factual conclusion based only
    on the provided documents, ensuring objectivity and factual accuracy. Reasoning
    and evidence in the actual output must be semantically equivalent to the human
    output, while ignoring stylistic differences. The judge also verifies that the
    model's answer is in the same language as the question and provides a brief justification
    for any major omissions, additions, mismatches, or failures to reference
    source documents. We also instructed the model to produce calibrated scores based
    on rubrics ranging from 0.0 "Factually incorrect" (0.0-0.3), "Mostly incorrect"
    (0.3-0.6), "Correct but missing minor details." (0.6-0.9), "100\% correct" (0.9-1.0),
    using the G-Eval implementation from the open-source DeepEval library\footnote{\url{https://github.com/confident-ai/deepeval}}.

    For the QA evaluation, we used \textit{Claude Sonnet 4}\footnote{anthropic.claude-sonnet-4-20250514-v1:0}
    with temperature set to 0.7 and \textit{top\_p} to 0.9.

    \subsection{Results and Discussions}
    \label{sec:results} Table \ref{tab:qa_eval_main_results} reports QA
    performance (G-Eval mean ± std) for each model and language. Models are grouped
    by sizes, small (<1B), medium (1-10B), and large (>10B), and score bands follow
    the rubrics in Section \ref{sec:llm_as_judge}. Higher scores indicate closer
    agreement with human reference answers.

    The results show a clear performance pattern across models and languages. Large
    instruction-tuned models obtain the highest mean scores in English and Spanish,
    placing them in the \textit{correct but missing minor details} range (0.6-0.9).
    Smaller 1B models fall substantially behind (0.15-0.35), often in the \textit{factually
    incorrect}, while bigger 3-8B variants place in the \textit{mostly correct}
    ranges (0.3-0.6). Beyond English and Spanish, cross-lingual performance
    degrades substantially for German, Italian, and Chinese, with even bigger models
    falling in the \textit{mostly correct} range (0.3-0.6), with high variance (std
    up to ~0.38). Notably, the reported standard deviations are non-negligible, indicating
    that this model's performance varies substantially across questions, and
    further studies are needed to identify this source of variability. Overall,
    while top-performing, large models achieve acceptable alignment with human answers
    in English and Spanish, there is still significant room for improvement in
    other languages to obtain better and more consistent cross-lingual
    performances. This baseline evaluation suggests that a careful selection of LLMs
    is crucial for QA tasks on résumés and JDs, especially in multilingual
    settings, and that further improvements are needed to ensure more robust HR applications.

    \section{Conclusion and Future Works}
    \label{sec:conclusion} We presented JobResQA, a multilingual benchmark
    designed to evaluate LLMs's machine reading comprehension capabilities on HR-specific
    tasks. The benchmark includes 581 QA pairs over 105 synthetic résumé-JD pairs
    across five languages, featuring three complexity levels: extractive fact-checking,
    multi-passage comprehension, and cross-document reasoning. By incorporating
    controlled demographic and professional attributes (via placeholders) and
    gender-inclusive design, JobResQA enables systematic fairness evaluation in HR
    applications. Our human-in-the-loop translation pipeline with MQM
    annotations demonstrates a quality-cost tradeoff in producing high-quality
    multilingual datasets. Baseline evaluations reveal substantial performance gaps
    across languages and model sizes, highlighting the need for improved
    multilingual capabilities in HR contexts, and further confirms the importance
    of open and shared data sets like JobResQA to promote research on this topic.

    \section{Limitations}
    \label{sec:limitations} We acknowledge the main limitations of our work:

    \paragraph{Localization and Translation Quality}
    Our automatic translations pipeline, despite employing human-in-the-loop feedback
    and review, may not fully capture the typical writing styles of real-world
    résumés and JDs from truly native speakers in each target language. Moreover,
    as noted in the error analysis, some inconsistencies in narrative voice (e.g.,
    first-person vs.\ third-person usage or infinitive constructions) may arise
    from paragraph-level machine translation of English source text and from the
    synthetic data generation process. In addition, the adoption of gender-inclusive
    rewriting, while linguistically and formally valid, may be less common in
    authentic résumés, slightly affecting the perceived naturalness of the text.
    These factors may influence tasks that rely on authentic linguistic patterns,
    and we plan to explore mitigation strategies in future work.

    \paragraph{QA Annotations Subjectivity}
    Annotators' judgments can be subjective, especially for complex questions
    that require cross-document reasoning, domain knowledge, or inference. This subjectivity
    may reduce alignment between LLM outputs and the human reference answers,
    thus negatively impacting the evaluation results.

    \section{Ethical Statement}
    \label{sec:ethical_statement}

    \paragraph{Implications for Bias Research in LLMs.}
    Our dataset is fully synthetic and anonymized, containing placeholder entities
    and gender-inclusive language. These design choices enable controlled
    investigation of potential bias attributes, such as demographic, gender, racial,
    and educational ones in LLMs applied to HR-related tasks. By systematically varying
    bias-related variables while removing personally identifiable information,
    our approach supports the study of model behavior based on content rather
    than identity cues.

    \paragraph{Potential Applications.}
    The dataset can facilitate the responsible development and evaluation of HR-oriented
    language technologies-such as chatbots or virtual assistants-for candidate screening,
    résumé parsing, and job-candidate matching. These high-risk applications
    should always be evaluated rigorously before being deployed in real-world
    settings, ensuring fairness, transparency, and accountability in decision-making.

    \paragraph{Human-in-the-Loop Annotations.}
    Professional annotators and native speakers were involved throughout all the
    stages of the dataset creation process. We ensured fair compensation and clear
    guidelines to support ethical labor practices and document generation and annotation
    processes for transparency and reproducibility. We believe and emphasize the
    role of human expertise to ensure high-quality data and produce ethical
    outcomes.

    \bibliography{references}
\end{document}

%% file: paul_tol_color_palette.tex
\usepackage[table,xcdraw]{xcolor} 

\definecolor{TolBrightBlue}{HTML}{4477AA}
\definecolor{TolBrightCyan}{HTML}{66CCEE}
\definecolor{TolBrightGreen}{HTML}{228833}
\definecolor{TolBrightYellow}{HTML}{CCBB44}
\definecolor{TolBrightRed}{HTML}{EE6677}
\definecolor{TolBrightPurple}{HTML}{AA3377}
\definecolor{TolBrightGrey}{HTML}{BBBBBB}

\definecolor{TolHighContrastBlue}{HTML}{004488}
\definecolor{TolHighContrastYellow}{HTML}{DDAA33}
\definecolor{TolHighContrastRed}{HTML}{BB5566}

\definecolor{TolVibrantOrange}{HTML}{EE7733}
\definecolor{TolVibrantBlue}{HTML}{0077BB}
\definecolor{TolVibrantCyan}{HTML}{33BBEE}
\definecolor{TolVibrantTeal}{HTML}{009988}
\definecolor{TolVibrantRed}{HTML}{CC3311}
\definecolor{TolVibrantMagenta}{HTML}{EE3377}
\definecolor{TolVibrantGrey}{HTML}{BBBBBB}

\definecolor{TolMutedCyan}{HTML}{88CCEE}
\definecolor{TolMutedTeal}{HTML}{44AA99}
\definecolor{TolMutedGreen}{HTML}{117733}
\definecolor{TolMutedBlue}{HTML}{332288}
\definecolor{TolMutedYellow}{HTML}{DDCC77}
\definecolor{TolMutedOlive}{HTML}{999933}
\definecolor{TolMutedPink}{HTML}{CC6677}
\definecolor{TolMutedWine}{HTML}{882255}
\definecolor{TolMutedViolet}{HTML}{AA4499}

\definecolor{TolPaleBlue}{HTML}{BBCCEE}
\definecolor{TolPaleCyan}{HTML}{CCEEFF}
\definecolor{TolPaleGreen}{HTML}{CCDDAA}
\definecolor{TolPaleYellow}{HTML}{EEEEBB}
\definecolor{TolPalePink}{HTML}{FFCCCC}
\definecolor{TolPaleGrey}{HTML}{DDDDDD}

\definecolor{TolSunset1}{HTML}{364B9A}
\definecolor{TolSunset2}{HTML}{4A7BB7}
\definecolor{TolSunset3}{HTML}{6EA6CD}
\definecolor{TolSunset4}{HTML}{98CAE1}
\definecolor{TolSunset5}{HTML}{C2E4EF}
\definecolor{TolSunset6}{HTML}{EAECCC}
\definecolor{TolSunset7}{HTML}{FEDA8B}
\definecolor{TolSunset8}{HTML}{FDB366}
\definecolor{TolSunset9}{HTML}{F67E4B}
\definecolor{TolSunset10}{HTML}{DD3D2D}
\definecolor{TolSunset11}{HTML}{A50026}

\definecolor{TolBuRd1}{HTML}{2166AC}
\definecolor{TolBuRd2}{HTML}{4393C3}
\definecolor{TolBuRd3}{HTML}{92C5DE}
\definecolor{TolBuRd4}{HTML}{D1E5F0}
\definecolor{TolBuRd5}{HTML}{F7F7F7}
\definecolor{TolBuRd6}{HTML}{FDDBC7}
\definecolor{TolBuRd7}{HTML}{F4A582}
\definecolor{TolBuRd8}{HTML}{D6604D}
\definecolor{TolBuRd9}{HTML}{B2182B}

\definecolor{TolPRGn1}{HTML}{762A83}
\definecolor{TolPRGn2}{HTML}{9970AB}
\definecolor{TolPRGn3}{HTML}{C2A5CF}
\definecolor{TolPRGn4}{HTML}{E7D4E8}
\definecolor{TolPRGn5}{HTML}{F7F7F7}
\definecolor{TolPRGn6}{HTML}{D9F0D3}
\definecolor{TolPRGn7}{HTML}{ACD39E}
\definecolor{TolPRGn8}{HTML}{5AAE61}
\definecolor{TolPRGn9}{HTML}{1B7837}

\definecolor{TolYlOrBr1}{HTML}{FFFFCC}
\definecolor{TolYlOrBr2}{HTML}{FFEDA0}
\definecolor{TolYlOrBr3}{HTML}{FED976}
\definecolor{TolYlOrBr4}{HTML}{FEB24C}
\definecolor{TolYlOrBr5}{HTML}{FD8D3C}
\definecolor{TolYlOrBr6}{HTML}{FC4E2A}
\definecolor{TolYlOrBr7}{HTML}{E31A1C}
\definecolor{TolYlOrBr8}{HTML}{BD0026}
\definecolor{TolYlOrBr9}{HTML}{800026}
